\documentclass[lettersize,journal]{IEEEtran}
\usepackage{amsmath,amsfonts,amssymb,amsthm}
\usepackage{algorithmic}
\usepackage{algorithm}
\usepackage{booktabs}
\usepackage{multicol}
\usepackage{multirow}
\usepackage{booktabs}
\usepackage{enumerate}
\usepackage{xcolor}
\usepackage{colortbl}
\usepackage{tabularray}
\usepackage{microtype}
\usepackage[utf8]{inputenc}
\usepackage[T1]{fontenc}
\UseTblrLibrary{booktabs}
\usepackage{tabularx}
\usepackage{hhline}
\usepackage{paralist}
\usepackage{marvosym}
\usepackage{array}
\usepackage[caption=false,font=scriptsize,labelfont=sf,textfont=sf]{subfig}
\usepackage{textcomp}
\usepackage{stfloats}
\usepackage{soul}
\usepackage{url}
\usepackage{verbatim}
\usepackage{graphicx}
\usepackage{hhline}
\usepackage{pifont}
\usepackage{cite}
\usepackage[colorlinks,
linkcolor=black,
anchorcolor=black,
citecolor=black]{hyperref}
\usepackage{orcidlink}
\def\etal{{\em et al.~}}

\newtheoremstyle{nonindentdef}
{5pt}   % 上方间距
{5pt}   % 下方间距
{\normalfont}  % 正文字体
{}      % 不缩进
{\bfseries} % 定理标题字体
{.}     % 后缀
{ }     % 标题和正文之间的空格
{\thmname{#1}\thmnumber{ #2}\thmnote{ (#3)}} % 定义标题格式

\theoremstyle{nonindentdef}

\begin{document}

\title{It’s Not the Target, It’s the Background: Rethinking Infrared Small Target Detection via Deep \\ Patch-Free Low-Rank Representations}

\author{Guoyi~Zhang~\orcidlink{0009-0004-2931-6698},~Guangsheng~Xu~\orcidlink{0009-0006-4235-9398},~Siyang~Chen~\orcidlink{0009-0008-9465-6647},~Han~Wang~\orcidlink{0009-0008-2794-2520}~and~Xiaohu~Zhang~\orcidlink{0000-0003-4907-1451}
	% <-this % stops a space
	\thanks{Manuscript received xxx, xxx; revised xxx, xxx.}
	\thanks{Corresponding authors: \emph{Han Wang and Xiaohu Zhang}}
	\thanks{Guoyi~Zhang, Guangsheng~Xu, Siyang~Chen, Han~Wang and Xiaohu~Zhang are with the School of Aeronautics and Astronautics, Sun Yat-sen University, Shenzhen 518107, Guangdong, China.(email:\href{mailto:zhanggy57@mail2.sysu.edu.cn}{\textcolor{black}{\{zhanggy57,xugsh6,chensy253,wangh737\}}}@mail2.sysu.edu.cn;\\zhangxiaohu@mail.sysu.edu.cn)}
}

% The paper headers
\markboth{Journal of \LaTeX\ Class Files,~Vol.~14, No.~8, August~2021}%
{Zhang \MakeLowercase{\textit{et al.}}: It’s Not the Target, It’s the Background: Rethinking Infrared Small Target Detection via Deep Patch-Free Low-Rank Representations}

%\IEEEpubid{0000--0000/00\$00.00~\copyright~2021 IEEE}
% Remember, if you use this you must call \IEEEpubidadjcol in the second
% column for its text to clear the IEEEpubid mark.

\maketitle

\begin{abstract}
\textcolor{blue}{This is the pre-acceptance version, to read the final version please go to \href{https://ieeexplore.ieee.org/document/11156113}{IEEE Transactions on Geoscience and Remote Sensing on IEEE Xplore}.} Infrared small target detection (IRSTD) remains a long-standing challenge in complex backgrounds due to low signal-to-clutter ratios (SCR), diverse target morphologies, and the absence of distinctive visual cues. While recent deep learning approaches aim to learn discriminative representations, the intrinsic variability and weak priors of small targets often lead to unstable performance.
In this paper, we propose a novel end-to-end IRSTD framework, termed LRRNet, which leverages the low-rank property of infrared image backgrounds. Inspired by the physical compressibility of cluttered scenes, our approach adopts a compression--reconstruction--subtraction (CRS) paradigm to directly model structure-aware low-rank background representations in the image domain, without relying on patch-based processing or explicit matrix decomposition.
To the best of our knowledge, this is the first work to directly learn low-rank background structures using deep neural networks in an end-to-end manner. Extensive experiments on multiple public datasets demonstrate that LRRNet outperforms 38 state-of-the-art methods in terms of detection accuracy, robustness, and computational efficiency. Remarkably, it achieves real-time performance with an average speed of 82.34 FPS. Evaluations on the challenging NoisySIRST dataset further confirm the model’s resilience to sensor noise. Code has been available at \href{https://github.com/HaLongbao/LRRNet/}{LRRNet}.
\end{abstract}

\begin{IEEEkeywords}
Infrared small target, low-rank matrix recovery, sparse representation, image segmentation, deep neural network.
\end{IEEEkeywords}

\section{Introduction}
\IEEEPARstart{I}{nfrared} small target detection (IRSTD) plays a vital role in reconnaissance \cite{chen2022local}, maritime surveillance \cite{strickland2023infrared}, and precision strike guidance \cite{yi2023spatial}, enabling long-range target acquisition.
However, detecting small targets in complex infrared scenes remains a significant challenge. This is primarily due to the presence of abundant clutter with target-like characteristics \cite{liu2023infrared}, the extremely small size and sparsity of targets \cite{LCRNet}, a severe foreground–background imbalance, low signal-to-clutter ratios (SCR) \cite{SeRankDet}, and the diverse, irregular shapes of true targets \cite{ISNet, SRNet, CSRNet}.

Generally, an infrared image $\mathbf{I}$ can be decomposed into the following components \cite{gao2013infrared}:
\begin{equation}
	\mathbf{I} = \mathbf{T} + \mathbf{B} + \mathbf{N}, \label{Eq:inf}
\end{equation}
where $\mathbf{T}$, $\mathbf{B}$ and $\mathbf{N}$ denote the target, background, and noise components, respectively. To address the inherent challenges, many existing methods incorporate prior knowledge of either the foreground or background into their design.

\begin{figure}[!t]
	\centering
	\includegraphics[width=\linewidth]{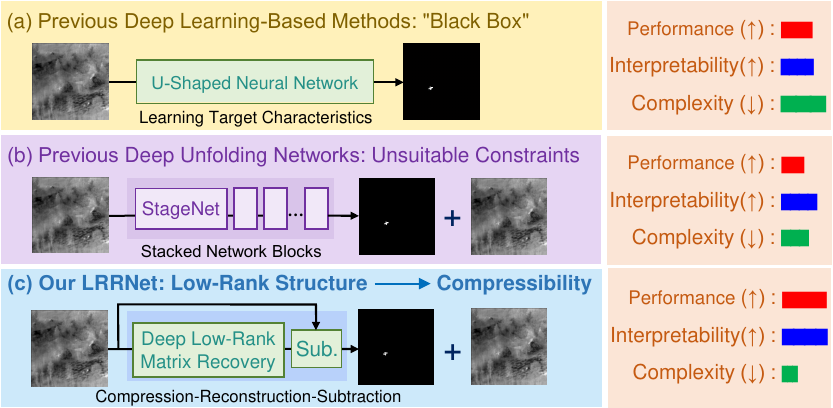}
	\caption{Comparison of the proposed LRRNet with existing data-driven methods.
		Unlike prior models that focus solely on target feature learning, LRRNet integrates image-domain low-rank priors without patch decomposition, enabling interpretable and efficient small-target detection.
		(a) Methods such as DNANet \cite{DNANet}, UIUNet \cite{UIUNet}, and SeRankDet \cite{SeRankDet} emphasize feature learning but struggle with limited robustness and interpretability due to target variability and the high complexity of deep-layer propagation.
		(b) RPCANet \cite{RPCANet} introduces interpretability via deep unfolding, yet imposes inappropriate constraints directly on the image domain instead of on patch-images, leading to suboptimal performance and computational inefficiency.
		(c) In contrast, our LRRNet adopts a patch-free compress–reconstruct–subtract paradigm:
		\ding{172} Small-target cues are retained in shallow layers, reducing both model depth and computational cost;
		\ding{173}  Low-rank priors enable robust modeling of cluttered backgrounds;
		\ding{174}  High-resolution subtraction isolates small targets effectively without the information loss associated with up/downsampling.}
	\label{fig:Compare}   
\end{figure}

The technical approach that emphasizes foreground characteristics typically models target sparsity. Traditional methods \cite{8705367,9822383,8245877} often rely on handcrafted features, but they frequently suffer from limited robustness and poor generalization across diverse scenarios \cite{liu2023infrared}.
To overcome these limitations, recent deep learning-based methods aim to capture intrinsic target attributes, including contrast \cite{Dai_2021_WACV, ALCNet}, shape \cite{ISNet, SRNet, CSRNet}, and sparsity \cite{SeRankDet, LCRNet}. These approaches have demonstrated notable improvements over conventional techniques in both accuracy and adaptability.
However, several challenges remain: (1) preserving small target information in deep layers often necessitates elaborate network architectures \cite{DNANet}, thereby increasing overall complexity \cite{MRF3Net}; (2) the absence of clear and discriminative features in small targets hinders robust representation learning \cite{LCRNet}; and (3) as inherently black-box models, deep networks typically lack interpretability, which limits their transparency and reliability in safety-critical applications \cite{RPCANet}.

\begin{figure*}[!t]
	\centering
	\includegraphics[width=1\linewidth]{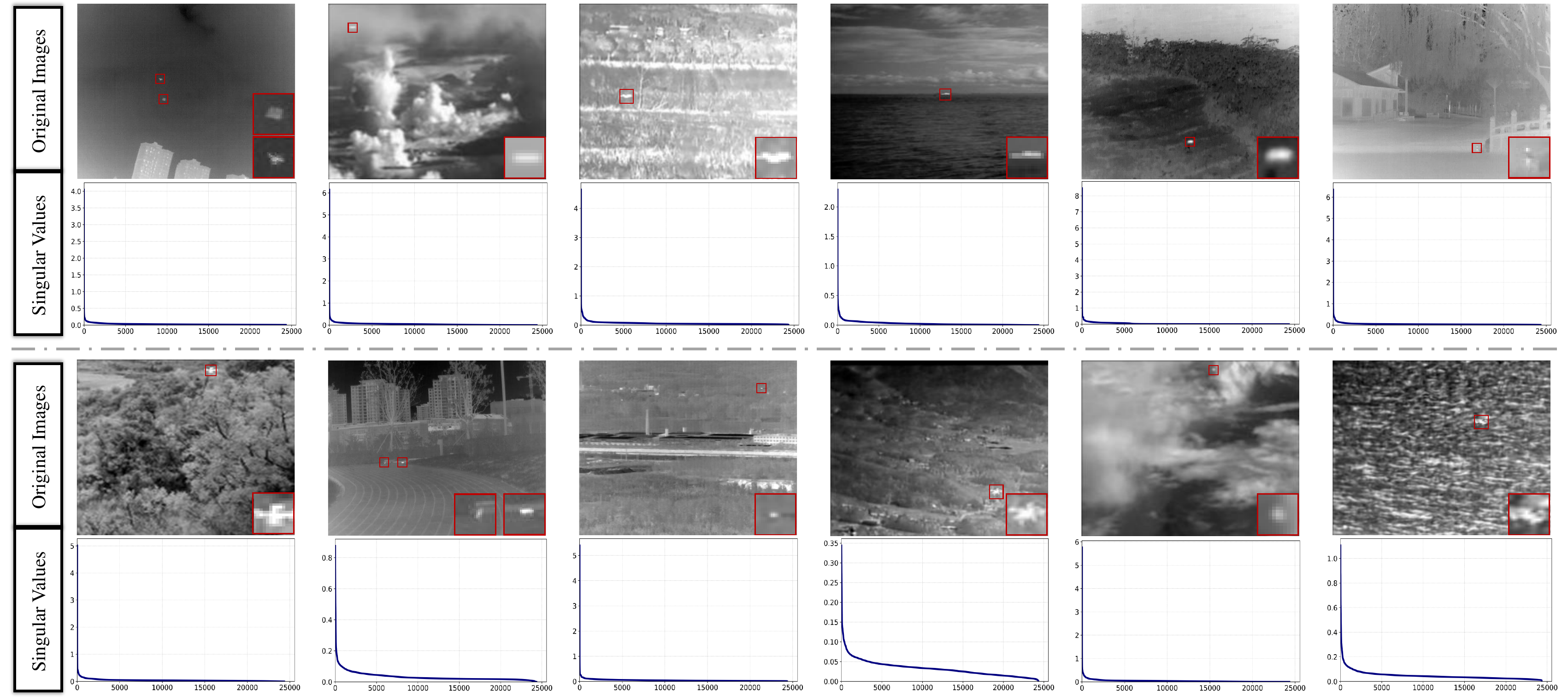}
	\caption{Important priors in infrared small target detection. The first row displays original infrared images from diverse scenes, and the second row shows the singular value distributions of their corresponding patch-image matrices \cite{gao2013infrared}. Three key observations can be made. \textbf{(1)} The targets exhibit significant variation in shape and size, and often appear with low contrast against cluttered backgrounds \cite{CSRNet}. This makes it difficult to robustly learn discriminative target-specific features \cite{LCRNet}. \textbf{(2)} Despite differences in clutter types, such as clouds, terrain, sea surface, sky, or man-made structures, the singular value curves demonstrate a consistent trend as the number of patches increases. This indicates that background regions share similar structural properties, which are more suitable for generalized modeling. \textbf{(3)} In highly complex scenes, the linear low-rank assumption becomes insufficient to represent the intrinsic structure of patch-images. This highlights the need for nonlinear low-rank constraints to more accurately capture the characteristics of real-world infrared backgrounds \cite{Wright-Ma-2022}.}
	\label{fig:Norm}
\end{figure*}

As shown in Fig.~\ref{fig:Norm}, infrared images often exhibit low-rank properties despite complex background clutter such as clouds, sea surfaces, or ground textures \cite{gao2013infrared, 9217948}, enabling effective background modeling. Compared to the unstable and diverse target features, the low-rank structure of backgrounds demonstrates strong robustness across different scenes. With a well-characterized background, dominant interference—primarily Gaussian noise \cite{10507231}—can be effectively suppressed, allowing sparse small target signals to be reliably recovered \cite{gao2013infrared}. This inherent robustness makes small target detection methods leveraging low-rank background priors highly effective. Such background–target decomposition is typically modeled using Robust Principal Component Analysis (RPCA) \cite{7172559,6180173,7994669} or its tensor extensions \cite{8908805,10354352,10078018}. However, recent work like RPCANet \cite{RPCANet} attempts to unfold RPCA into a trainable network but imposes low-rank constraints on patch-images rather than on the entire image \cite{10975043}, which results in suboptimal detection performance and significant computational overhead, especially on high-resolution infrared data.

Motivated by these limitations, we propose a patch-free Low-Rank Representation Network (LRRNet) that integrates low-rank constraints directly in the image domain, eliminating the need for patch-image construction. This design is based on \textbf{a key insight}: the low-rank nature of infrared images implies an inherent compressibility---when the rank is sufficiently small, the image information can be sparsely encoded while being nearly ``losslessly'' preserved.

To this end, as illustrated in Fig.~\ref{fig:Compare}, we introduce a compact and effective compress--reconstruct--subtract paradigm. Unlike conventional deep networks that propagate small-target features through deep layers, often requiring elaborate architectural design, LRRNet preserves small-target cues in high-resolution feature maps, avoiding information degradation caused by repeated downsampling and upsampling. Furthermore, the reconstruction process, inspired by the compressibility of low-rank signals, enhances both the interpretability and efficiency of the proposed model, while avoiding the pitfalls of deep propagation or memory-heavy patch construction.

We summarize the main contributions of the paper as follows:
\begin{itemize}
	\item We propose LRRNet, the first end-to-end deep network that directly models low-rank background priors in the image domain, without relying on handcrafted patch-based matrices or explicit decomposition.
	\item We introduce compression--reconstruction--subtraction (CRS) paradigm that leverages the physical interpretation of low-rank structures in infrared backgrounds. This design preserves small-target cues in shallow layers, avoiding the need for deep propagation and eliminating the complexity of excessively deep architectures.
	\item We conduct extensive experiments on multiple public benchmarks, comparing LRRNet with over 38 state-of-the-art methods. Results demonstrate superior accuracy and efficiency. In addition, evaluations on the NoisySIRST dataset \cite{SeRankDet} highlight the model’s strong robustness against sensor-induced noise.
\end{itemize}

The remainder of this paper is organized as follows. Section~\ref{Section:Related_Work} provides a brief review of related work. Section~\ref{Section:Preliminaries} discusses several mathematical foundations of low-rank and sparse decomposition. Section~\ref{Section:ProposedMethod} presents the proposed method in detail. Experimental results and corresponding analyses are reported in Section~\ref{Section:Experiment}. Finally, Section~\ref{Section:Conclusion} concludes the paper.

\section{Related Work} \label{Section:Related_Work}
We briefly review related works from three aspects: infrared small target detection, low-rank representation and learned image compression model.
\subsection{Infrared Small Target Detection}
Infrared small target detection algorithms are generally divided into two categories: those that model target features based on structural sparsity, and those that focus on learning background features with low-rank structure.

In the approach of learning target structural sparsity, early model-driven algorithms relied on handcrafted features and target priors specific to scenarios like contrast and Gaussian-like grayscale distributions, which limited their generalization and robustness in complex scenes \cite{8705367,9822383,8245877,chen2022local}. To overcome these challenges, data-driven methods were introduced, incorporating small target information into deep networks to learn discriminative features \cite{SeRankDet}, achieving superior performance. Methods such as ACMNet \cite{Dai_2021_WACV}, ALCNet \cite{ALCNet}, and RDIAN \cite{RDIAN} use contrast constraints for efficient target information transmission, being parameter-efficient and fast, but suffer from high false alarm rates in complex scenes. DNANet \cite{DNANet} and UIUNet \cite{UIUNet} preserve small target information through dense nesting, but their inference speed is very slow. ISNet \cite{ISNet}, SRNet \cite{SRNet}, and CSRNet \cite{CSRNet} leverage shape bias, but shape priors are more effective for large targets, making it difficult to distinguish point-like targets from false alarms. Gao \etal \cite{liu2023infrared} introduced ViT for infrared small target detection, utilizing long-range feature correlations to differentiate targets and false alarms. However, self-attention’s high complexity limits its ability to handle high-resolution data, and ViT’s lack of inductive bias causes small target information loss, requiring hybrid architectures with added complexity \cite{MTUNet}. SeRankDet \cite{SeRankDet} uses Top-K attention to enforce sparsity on the target but suffers from severe over-parameterization (108M parameters). LCRNet \cite{LCRNet} relaxes sparsity constraints and employs large kernel attention to construct dynamic local contexts, achieving excellent performance with only 1.6M parameters. However, it still requires incorporating target information into deep networks, and the lack of support for large kernels in PyTorch leads to slow inference speeds.

The approach to modeling the background’s low-rank structure primarily involves model-driven methods \cite{gao2013infrared, dai2017non, dai2017reweighted, zhang2018infrared, zhang2019infrared, zhang2019infrared1}, which transform the original small target detection problem into RPCA \cite{7172559,6180173,7994669} or TensorRPCA \cite{8908805,10354352,10078018} problems by constructing patch-images \cite{gao2013infrared}. To further enhance the separability of the target and background, scene-specific regularization terms are often introduced \cite{yi2023spatial}, simultaneously constraining the target's structural sparsity. However, such methods struggle to model complex real-world scenes, as they capture only specific structures and lack the ability to understand semantic information, leading to poor performance in complex scenarios \cite{liu2023infrared}.

In contrast to the aforementioned methods, we propose a patch-free Low-Rank Representation Network (LRRNet) that directly incorporates low-rank constraints in the image domain, eliminating the need for patch-image construction. This design is based on a key insight: the low-rank nature of infrared images suggests inherent compressibility—when the rank is sufficiently small, the image information can be sparsely encoded while nearly "losslessly" preserved. This approach retains small target cues in shallow layers, avoiding deep propagation and eliminating the complexity associated with excessively deep architectures.

\subsection{Low-Rank Representation}

Low-rank representation \cite{7172559} has long served as a fundamental tool and theoretical foundation for analyzing high-dimensional data exhibiting underlying low-dimensional structures \cite{Wright-Ma-2022}. Over the past decades, it has played a pivotal role in various domains, including target detection \cite{6180173}, data recovery \cite{8908805}, and computational imaging \cite{10354352}. Early approaches commonly relied on carefully designed mathematical models incorporating various constraints \cite{10078018}, such as noise assumptions \cite{7994669}, to solve the underlying problems. However, these models are highly nonlinear and suffer from significant computational inefficiency as problem scale grows \cite{10487002}. Moreover, handcrafted heuristic constraints often exhibit limited generalizability to the diverse and complex challenges posed by real-world scenarios \cite{RPCANet}.

While several recent deep learning methods \cite{10601492,8620376,10684569} have aimed to capture low-rank structures in data, they generally operate on inputs that inherently satisfy low-rank constraints. In infrared small target detection, this typically requires constructing patch-based image representations \cite{gao2013infrared} prior to processing the resulting data matrices. For example, for a $256\times200$ image, the classical IPI method \cite{gao2013infrared} generates a patch-image data matrix sized $2500 \times 336$. For larger images such as $512 \times 512$ in the IRSTD-1K dataset \cite{ISNet}, this transformation produces an excessively large matrix, imposing substantial GPU memory demands.
In contrast, our approach employs a compression–reconstruction–subtraction (CRS) paradigm to directly model structure-aware low-rank background representations in the image domain, obviating the need for patch-based processing or explicit matrix decomposition.

\subsection{Learned Image Compression Model}
Although motivated by the low-rank property of infrared images, which implies intrinsic compressibility, where a sufficiently low-rank image can be sparsely encoded while nearly preserving information without loss—our proposed compression–reconstruction–subtraction paradigm fundamentally \textbf{differs} from conventional learned image compression methods \cite{xiang2024remote} in low-level vision tasks.

Typical learned image compression approaches aim to minimize storage and transmission costs under the constraint of maintaining perceptual image quality, generally without making assumptions about specific image content or focusing on the compression of particular semantic features \cite{li2021learning}. In contrast, our method is designed to reconstruct the background in infrared images corrupted by Gaussian noise and structured sparse signals representing small targets \cite{gao2013infrared}. Moreover, due to the properties of small targets, such as local smoothness and energy concentration \cite{Dai_2021_WACV}, which differentiate them from the smoothly varying background and Gaussian noise, our compression-reconstruction framework does \textbf{not} require highly accurate background recovery. Instead, it identifies the correlation between false alarm sources and the background during the reconstruction process, thereby effectively suppressing the impact of false alarms.

\section{Preliminaries} \label{Section:Preliminaries}
In this section, we discuss some mathematical theories of low-rank and sparse decomposition to further enhance the interpretability of our method.

Let us reformulate Eq. (\ref{Eq:inf}) into a decomposition form of the data matrix constructed from infrared patch-images, which is given as follows:
\begin{equation}
	\mathbf{D} = \mathbf{T} + \mathbf{B} + \mathbf{N}, 
\end{equation}
the explicit constraints of the low-rank and sparse decomposition are formulated as follows:
\begin{equation}
	\setlength{\abovedisplayskip}{4pt}
	\min \limits_{\mathbf{B},\mathbf{T}} rank(\mathbf{B}) + \lambda \left\| \mathbf{T} \right\|_0 \quad s.t.~\mathbf{D} = \mathbf{B} + \mathbf{T} \enspace,
	\vspace{-0.2cm}
	\label{RPCA}
\end{equation}
where $\lambda$ indicates a positive trade-off parameter, and ${\left\|  \cdot  \right\|_0}$ demotes the $l_0$-norm as the number of nonzero entries.

In this case, the deep unfolding framework involves alternately solving Eq. (\ref{RPCA}) via the ADMM algorithm, and subsequently transforming each iteration into a specific layer within the network.
However, since the explicit low-rank constraint is imposed on patch-images rather than directly on the image domain, unfolding the model in the image domain may introduce inappropriate constraints. Once the patch-images are constructed, the resulting data matrix becomes extremely large. For example, a single $512\times512$ image yields a 
$2500\times2209$ matrix under the IPI \cite{gao2013infrared} construction scheme. Moreover, as low-rankness is inherently a global property, it cannot be effectively enforced through conventional block-wise processing.

In contrast to deep unfolding methods, our core insight is that the low-rank property of image backgrounds is an inherent structural characteristic, and explicit matrix decomposition is merely one specific way to exploit this property—by no means the only one. Therefore, we interpret background low-rankness as a form of compressibility, aligning with the information-theoretic principle \cite{Wright-Ma-2022} that ``\textit{learning is compression, and compression is learning.}'' Based on this perspective, we propose a compression–reconstruction–subtraction pipeline to enable small object detection.

\section{Proposed Method} \label{Section:ProposedMethod}
\subsection{Overview}
\begin{figure*}[!t]
	\centering
	\includegraphics[width=1\linewidth]{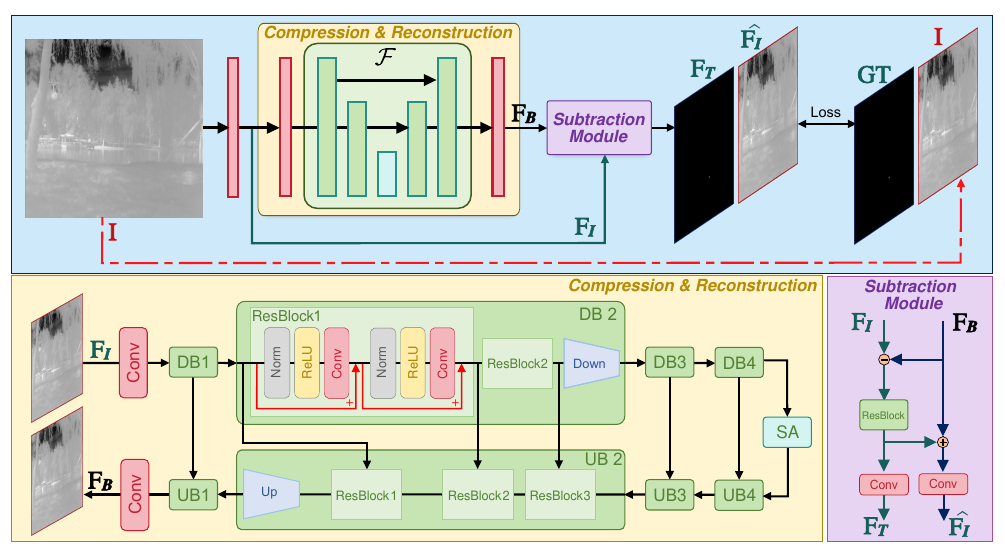}
	\caption{Overview of the proposed LRRNet architecture, which follows the compression–reconstruction–subtraction paradigm. The network learns structure-aware low-rank background representations through a densely connected encoder–decoder backbone, enabling effective small target extraction by subtracting the reconstructed background from the input features. DB and UB denote downsampling and upsampling blocks at each stage, respectively, each consisting of a residual block followed by a downsampling or upsampling convolution. SA represents a self-attention module used to enhance feature representation. The entire network is jointly optimized by a segmentation loss and a reconstruction loss.}
	\label{fig:arch}
\end{figure*}

The overall structure of the proposed method is illustrated in Fig.~\ref{fig:arch}. It follows the compression–reconstruction–subtraction paradigm, where a densely connected U-Net is employed to compress and reconstruct the background. Small target features are effectively extracted by subtracting the reconstructed background from the input features. The subtraction module isolates target-relevant information while combining the reconstructed background with extracted target features to approximate the original input. The entire network is jointly optimized using both segmentation loss and reconstruction loss, ensuring accurate target localization while preserving structural consistency.

\subsection{Compression and Reconstruction Module}
The overall structure of the proposed Compression and Reconstruction Module is illustrated in Fig.~\ref{fig:arch}. This module serves as the core component of our framework. It captures the low-rank structure of infrared image backgrounds through a compression–reconstruction process, without relying on patch-wise construction or explicit matrix decomposition, while remaining structure-aware.

\noindent\textbf{Motivation}. The proposed Compression and Reconstruction Module is driven by three key insights:  
(1) The low-rank nature of infrared image backgrounds implies that they can be effectively compressed \cite{Wright-Ma-2022}, and such compression can be efficiently achieved using a UNet-based architecture \cite{li2021learning};  
(2) To enhance noise robustness, reduce computational complexity, and exploit long-range dependencies for improved compression efficiency, we introduce a self-attention mechanism \cite{liu2023infrared} at the lowest-resolution feature map;  
(3) Although residual connections are effective in propagating information through the network, they may lead to feature confusion \cite{zhou2021deepvit}, which is particularly detrimental for accurately separating background signals. To address this, we employ densely connected long skip connections between encoder and decoder blocks at the same resolution level.

\noindent\textbf{Overview of the Proposed Compression and Reconstruction Module}.  
The proposed compression-reconstruction module can be viewed as a specialized mapping \( \mathcal{F}: \mathbf{F}_I \rightarrow \mathbf{F}_B \), where \(\mathbf{F}_I \in\mathbb{R}^{W\times H \times C}\) denotes the feature map extracted from the original image, and \(\mathbf{F}_B\) represents the corresponding reconstructed background feature map. Notably, \(\mathbf{F}_B\) shares the same spatial and channel dimensions as \(\mathbf{F}_I\).
Given an input feature map $\mathbf{I} \in\mathbb{R}^{W\times H \times 1}$, we first apply a $3 \times 3$ convolution to suppress local noise. 
\begin{equation}
	\mathbf{F_\mathit{I}} = \mathrm{Conv}_{3\times3}(\mathbf{I}).
\end{equation}
The processed features are then passed through successive stages of the densely connected U-Net. At the lowest resolution level, a self-attention mechanism \cite{liu2023infrared} is employed to capture long-range dependencies between features.

Each stage is composed of residual blocks \cite{huang2023revisiting}, and densely connected long skip connections are established between encoder and decoder blocks at the same resolution level. For an input feature map $\mathbf{X} \in\mathbb{R}^{W\times H \times C}$, the processing flow within each residual block \cite{huang2023revisiting} is as follows: 
\begin{equation}
	\mathbf{X}^{\prime}=\mathbf{X}+\lambda_1\left(\mathrm{Norm}_1(\mathrm{ReLU}(\mathrm{Conv}_{3\times3}(\mathbf{X})))\right),
\end{equation}
\begin{equation}
	\mathbf{X}^{\prime\prime}=\mathbf{X}^{\prime}+\lambda_2\left(\mathrm{Norm}_2(\mathrm{ReLU}(\mathrm{Conv}_{3\times3}(\mathbf{X}^{\prime})))\right),
\end{equation}
where $\mathrm{Norm}_1(\cdot)$ denotes Group Normalization \cite{wu2018group}, $\lambda_1$ and $\lambda_2$ are learnable parameters that define the residual weighting in the LayerScale \cite{touvron2021going} mechanism. We adopt the pre-activation ResBlock \cite{huang2023revisiting} design due to its improved robustness to noise \cite{huang2023revisiting}, which is particularly critical for accurate background compression and reconstruction under noisy conditions.

At the lowest resolution level, a standard query-key-value (qkv) self-attention mechanism \cite{dosovitskiy2020image} is applied. Since the feature map $\mathbf{X}$ has undergone four downsampling operations, its spatial resolution is reduced from the original $H\times W$ to $\frac{H}{16} \times \frac{W}{16}$ resulting in negligible computational overhead. The standard qkv self-attention is formulated as:
\begin{equation}
	\text{Attention}(\mathbf{Q}, \mathbf{K}, \mathbf{V}) = \text{Softmax}\left(\frac{\mathbf{QK}^\top}{\sqrt{d_k}}\right)\mathbf{V},
\end{equation}
where $\mathbf{Q} = \mathbf{X}\mathbf{W}^Q$, $\mathbf{K} = \mathbf{X}\mathbf{W}^K$, and $\mathbf{V} = \mathbf{X}\mathbf{W}^V$ are the query, key, and value projections of the input feature map $\mathbf{X}$, and $d_k$ denotes the dimensionality of the key vectors used for scaling. Recent studies \cite{yun2024shvit} have shown that when the number of channels is large, the information captured across different channels tends to be highly redundant. As a result, employing multi-head attention in such scenarios provides limited benefit. In our design, self-attention is applied at the lowest resolution level, where the channel dimension is the largest within the network. Therefore, we adopt single-head attention instead of a multi-head formulation to reduce redundancy and computational overhead.

\noindent\textbf{Discussion}.  
Our Compression and Reconstruction Module can be \textit{\textbf{interpreted as a nonlinearized Multiscale Hybrid Model (MHM)}}~\cite{hong2006multiscale}, offering a degree of physical interpretability to the compression–reconstruction process. Recent studies~\cite{falck2022multi, williams2023unified} have shown that U-Net architectures possess inherent regularization properties and share a close connection with wavelet representations. 
Building on this insight, our densely connected U-Net extends this capacity by constructing multiscale, \textit{\textbf{multi-subspace}} feature representations with fine-grained granularity. The nonlinearity of neural networks further enables flexible, nonlinear mixing of these subspaces, making the model more adaptable to complex real-world data. Therefore, our module can be naturally regarded as a nonlinear extension of the original MHM~\cite{hong2006multiscale}, preserving its multiscale formulation while enhancing its representational capacity. MHMs~\cite{hu2014multiresolution} have proven effective in various tasks, including image compression and reconstruction, supporting the relevance of this interpretation.

\subsection{Subtraction Module}
The overall structure of the subtraction module is illustrated in Fig. \ref{fig:arch}. Its core objective is to isolate small target information by leveraging the reconstructed low-rank background and the corresponding feature maps from the original input image. This design enables the effective decoupling of target-relevant sparse components from the background, facilitating precise localization in cluttered scenes.

\noindent\textbf{Motivation}. The design of the Subtraction Module is driven by the following two insights: (1) Small target information predominantly resides in high-resolution feature maps \cite{ilnet}. By subtracting the reconstructed background features from the original ones, the target can be effectively isolated. This approach avoids multiple downsampling operations that often distort target shapes \cite{ISNet} and lead to over-segmentation \cite{liu2023infrared}.
(2) After background removal, the residual features primarily contain small targets and Gaussian noise \cite{gao2013infrared}. Owing to the local smoothness of small targets, distinguishing them from noise becomes relatively straightforward, eliminating the need for complex network structures \cite{8365806}.

\noindent\textbf{Overview of the proposed subtraction module}. Given two input feature maps, $\mathbf{F_\mathit{I}} \in\mathbb{R}^{W\times H \times C}$ and $\mathbf{F_\mathit{B}} \in\mathbb{R}^{W\times H \times C}$, the target feature map $\mathbf{F_\mathit{T}} \in\mathbb{R}^{W\times H \times C}$ is computed as follows:
\begin{equation}
	\mathbf{F_\mathit{T}} = \text{ResBlock}(\mathbf{F_\mathit{I}} - \mathbf{F_\mathit{B}}).
\end{equation}

Subsequently, the reconstructed feature map 
$\hat{\mathbf{F_\mathit{I}}}$ is obtained by adding the background and target feature maps:
\begin{equation}
	\hat{\mathbf{F_\mathit{I}}} = \mathbf{F_\mathit{T}} + \mathbf{F_\mathit{B}}.
\end{equation}

Finally, both $\mathbf{F_\mathit{T}}$ and $\mathbf{F_\mathit{I}}$ are downsampled through a 3×3 convolution to reduce the channel dimensions and output accordingly.

\noindent\textbf{Discussion}. Despite its simplicity, the proposed subtraction module offers several compelling advantages: (1) Since the success of DnCNN \cite{7839189}, the effectiveness of residual blocks for denoising tasks has been extensively validated both theoretically and empirically \cite{8365806}.  (2) Applying a reconstruction loss to the fused feature $\hat{\mathbf{F_\mathit{I}}}$ effectively facilitates the compression–reconstruction module in learning the compressible nature of the background \cite{10844040}. Moreover, as the noise in $\hat{\mathbf{F_\mathit{I}}}$ primarily originates from the target feature map $\mathbf{F_\mathit{T}}$, an appropriately designed loss function can further suppress the residual noise in both $\hat{\mathbf{F_\mathit{I}}}$ and $\mathbf{F_\mathit{T}}$, thereby improving the overall detection robustness.

\subsection{Loss Function}
The loss function in our framework consists of two main components: a segmentation loss for supervising small target localization and a reconstruction loss for constraining the background compression-reconstruction process.

For small target segmentation, considering the inherent imbalance between the target and the background \cite{liu2023infrared}, we adopt the SoftIoU loss to measure the similarity between the predicted confidence map and the ground truth label \cite{SeRankDet}. Furthermore, to ensure that the optimization process is properly guided even in the absence of overlap between the predicted and ground truth labels \cite{MSHNet}, we incorporate L1 Loss, which serves as a more robust constraint. Consequently, the final segmentation loss is composed of both the SoftIoU loss and the L1 loss.

The reconstruction loss is designed with two primary objectives: (1) to provide effective supervision for learning the compressibility of the background, and (2) to suppress the Gaussian noise inherent in the reconstructed features. To this end, we employ the Mean Squared Error (MSE) loss, which is a standard and theoretically optimal choice for modeling Gaussian white noise \cite{Wright-Ma-2022}. Due to MSE loss being computed over the entire image, it may overlook small, critical details, particularly for small targets or subtle features. Hence, reconstruction supervision leverages the original input image \textbf{rather than} requiring ground truth background data.

The total loss is defined as a weighted sum of the segmentation and reconstruction losses:
\begin{equation}
	\mathcal{L}_{\text{total}} = \mathcal{L}_{\text{seg}} + \lambda \mathcal{L}_{\text{rec}},
\end{equation}
where $\mathcal{L}_{\text{seg}}$ denotes the combined SoftIoU and L1 loss for small target segmentation, $\mathcal{L}_{\text{rec}}$ represents the MSE loss for background reconstruction, and $\lambda = 0.1$ is a weighting factor that balances the two objectives.

\section{Experiment}\label{Section:Experiment}
\subsection{Experimental Setup}
\subsubsection{Datasets} We evaluate the proposed method on three widely used datasets: IRSTD-1K \cite{ISNet}, SIRSTAUG \cite{AGPCNet}, and NUDT-SIRST \cite{DNANet}. These diverse and challenging benchmarks \cite{SeRankDet} provide a rigorous testbed for assessing performance and generalization. The three datasets exhibit markedly different target size distributions. IRSTD-1K demonstrates strong multi-scale characteristics, with 18\% of targets in the range (0, 10], 46\% in (10, 40], 27\% in (40, 100], and 9\% above 100 pixels. SIRSTAUG is dominated by typical small targets, with 97\% in the (0, 10] range. NUDT-SIRST presents cross-scale characteristics, with 25\% of targets in (0, 10], 44\% in (10, 40], and 30\% in (40, 100], but no particularly large targets. This diversity allows for a comprehensive evaluation of the algorithm’s adaptability across varying target scales.

\subsubsection{Evaluation Metrics} In this study, four widely used evaluation metrics including the probability of detection ($P_d$), false alarm rate ($F_a$), intersection over union (IoU) and normalized intersection over union (nIoU) \cite{Dai_2021_WACV}, are used for performance evaluation. Among them, the $P_d$ and $F_a$ are object-level metrics, which are defined as follows:
\begin{equation}
	P_d=\frac{TP}{TP + FN},
\end{equation}
\begin{equation}
	F_a=\frac{FP}{FP + TN}.
\end{equation}
The IoU and nIoU \cite{Dai_2021_WACV} are pixel-level metrics, which are defined as follows:
\begin{equation}
	IoU=\frac{TP}{T+P-TP},
\end{equation}
\begin{equation}
	nIoU=\frac{1}{N}\sum_i^N\frac{TP(i)}{T(i)+P(i)-TP(i)}.
\end{equation}
where $N$ is the total number of samples, $T$, $P$, $FP$ and $TP$ denote the number of ground truth, predicted positive pixels, the number of false positive, and the number of true positive respectively.

\subsubsection{Implementation Details} The proposed method is implemented in PyTorch 1.8.2 and accelerated with CUDA 11.2. The network is trained using the Adam optimizer with an initial learning rate of 1e-4, betas (0.9, 0.999), eps 1e-8, weight decay 0 on an NVIDIA A100 GPU. Learning rate decay follows the poly policy \cite{RPCANet}, with a batch size of 8 and a maximum of 400 epochs. To ensure reproducibility, no data augmentation is used, and the input is a single-channel image normalized by dividing grayscale values by 255.0.

\subsection{Comparison with State-of-the-Arts}
\begin{table*}
	\renewcommand\arraystretch{1.2}
	\footnotesize
	\centering
	\caption{Comparison with Other State-of-the-art methods on three datasets. \textbf{\textcolor{cyan}{Blue}}, \textbf{\textcolor{orange}{orange}}, and \textbf{\textcolor{green}{green}} represent the best, second-best, and third-best metrics in each column, respectively. The metrics considered include IoU ($10^{-2}$), $\text{nIoU}$ ($10^{-2}$), $P_d$ ($10^{-2}$), $F_a$ ($10^{-6}$). }
	\label{tab:sota}
	% \vspace{-1\baselineskip}
	\setlength{\tabcolsep}{3.5pt}
	\begin{tabular}{l|c|c|c|cccc|cccc|cccc}
		\noalign{\hrule height 1pt}
		\multirow{2}{*}{Method} & \multirow{2}{*}{Publish} & \multirow{2}{*}{Param} & \multirow{2}{*}{FLOPs} & \multicolumn{4}{c|}{IRSTD-1k}   & \multicolumn{4}{c|}{SIRSTAUG}   & \multicolumn{4}{c}{NUDT-SIRST} \\ \multicolumn{1}{l|}{} & \multicolumn{1}{l|}{}  &\multicolumn{1}{l|}{} & \multicolumn{1}{l|}{} 
		& IoU $\uparrow$ &  nIoU $\uparrow$ & $P_d$ $\uparrow$ & $F_a$ $\downarrow$ & IoU $\uparrow$ &  nIoU $\uparrow$ & $P_d$ $\uparrow$ & $F_a$ $\downarrow$ & IoU $\uparrow$ &  nIoU $\uparrow$ & $P_d$ $\uparrow$ & $F_a$ $\downarrow$ \\
		\noalign{\hrule height 1pt}
		\multicolumn{16}{l}{\textit{Low-rank and Sparse Decomposition}}  \\
		\hline
		IPI \cite{gao2013infrared}   & TIP'13    & $\textemdash$ & $\textemdash$ & 27.92 & 30.12 & 81.37 & 16.18 & 25.16 & 34.64 & 76.26 & 43.41 & 28.63 & 38.18 & 74.49 & 41.23 \\
		NIPPS \cite{dai2017non}  & IPT'17    & $\textemdash$ & $\textemdash$ & 5.424 & 8.52  & 65.47 & 31.59 & 17.03 & 27.54 & 76.88 & 52.01 & 30.21 & 40.94 & 89.41 & 35.18 \\
		RIPT \cite{dai2017reweighted}  & JSTARS'17 & $\textemdash$ & $\textemdash$ & 14.11 & 17.43 & 77.55 & 28.31 & 24.13 & 33.98 & 78.54 & 56.24 & 29.17 & 36.12 & 91.85 & 344.3 \\
		NRAM \cite{zhang2018infrared}  & RS'18     & $\textemdash$ & $\textemdash$ & 9.882 & 18.71 & 72.48 & 24.73 & 8.972 & 15.27 & 71.47 & 68.98 & 12.08 & 18.61 & 72.58 & 84.77 \\
		NOLC \cite{zhang2019infrared1}  & RS'19     & $\textemdash$ & $\textemdash$ & 12.39 & 22.18 & 75.38 & 21.94 & 12.67 & 20.87 & 74.66 & 67.31 & 23.87 & 34.9  & 85.47 & 58.2  \\
		PSTNN \cite{zhang2019infrared} & RS'19     & $\textemdash$ & $\textemdash$ & 24.57 & 28.71 & 71.99 & 35.26 & 19.14 & 27.16 & 73.14 & 61.58 & 27.72 & 39.8  & 66.13 & 44.17 \\
		\hline
		\multicolumn{16}{l}{\textit{Deep Learning Methods: Learning Target Characteristics}}  \\
		\hline
		MDvsFA \cite{wang2019miss}                  & ICCV'19  & 3.919M                         & 998.6G  & 49.50                     & 47.41  & 82.11                     & 80.33 &  $\textemdash$ &  $\textemdash$ &  $\textemdash$ &  $\textemdash$ & 75.14 &  $\textemdash$ & 90.47 & 25.34 \\
		ACM  \cite{Dai_2021_WACV}                     & WACV'21 & 0.5198M                        & 2.009G  & 63.39                     & 60.81  & 91.25                     & 8.961 & 73.84  & 69.83  & 97.52  & 76.35  & 68.48                     & 69.26  & 96.26  & 10.27  \\
		ALCNet  \cite{ALCNet}                  & TGRS'21 & 0.5404M                        & 2.127G  & 62.05                     & 59.58  & 92.19                     & 31.56 &  $\textemdash$ &  $\textemdash$ &  $\textemdash$ &  $\textemdash$ & 61.13   & 60.61  & 96.51 & 29.09 \\
		FC3-Net  \cite{FC3-Net}                 & MM'22   & 6.896M                         & 10.57G  & 64.98                     & 63.59  & 92.93                     & 15.73 & $\textemdash$ & $\textemdash$ & $\textemdash$ & $\textemdash$ & $\textemdash$                    & $\textemdash$ & $\textemdash$ & $\textemdash$ \\
		DNANet \cite{DNANet}                   & TIP'22  & 4.6968M                        & 56.08G  & 68.87                     & 67.53  & 94.95                     & 13.38 &  \textbf{\textcolor{green}{74.88}}  & 70.23  & 97.80   & 30.07  & 92.67                     & 92.09  & \textbf{\textcolor{orange}{99.53}}  & 2.347  \\
		ISNet \cite{ISNet}                    & CVPR'22 & 1.09M                          & 121.90G & 68.77                     & 64.84  & 95.56                     & 15.39 & 72.50   & 70.84  & 98.41  & 28.61  & 89.81                     & 88.92  & 99.12  & 4.211  \\
		AGPCNet \cite{AGPCNet}                  & TAES'23 & 12.36M                         & 327.54G & 68.81                     & 66.18  & 94.26                     & 15.85 & 74.71  & \textbf{\textcolor{orange}{71.49}}  & 97.67  & 34.84  & 88.71                     & 87.48  & 97.57  & 7.541  \\
		UIUNet  \cite{UIUNet}                  & TIP'23  & 50.54M                         & 217.85G & 69.13                     & 67.19  & 94.27                     & 16.47 & 74.24  & 70.57  & 98.35  & \textbf{\textcolor{green}{23.13}}  & 90.77                     & 90.17  & 99.29  & 2.39   \\
		RDIAN  \cite{RDIAN}                   & TGRS'23 & 0.131M                         & 14.76G  & 64.37                     & 64.90   & 92.26                     & 18.2  & 74.19  & 69.8   & \textbf{\textcolor{cyan}{99.17}}  & 23.97  & 81.06                     & 81.72  & 98.36  & 14.09  \\
		MTUNet \cite{MTUNet}                   & TGRS'23 & 4.07M                          & 24.43G  & 67.50                      & 66.15  & 93.27                     & 14.75 & 74.70   & 70.66  & \textbf{\textcolor{green}{98.49}}  & 39.73  & 87.49                     & 87.70   & 98.60   & 3.76   \\
		RepISD-Net \cite{wu2023repisd}                  & TGRS'23 & 0.28M                          & 25.76G  & 65.45                     & $\textemdash$  & 91.59                     & 7.62 & $\textemdash$   & $\textemdash$  & $\textemdash$  & $\textemdash$  & 89.44                     & $\textemdash$   & 98.65   & 6.18   \\
		SRNet  \cite{SRNet}                   & TMM'23  & 0.4045M & $\textemdash$  & 69.45                     & 65.51  & \textbf{\textcolor{orange}{96.77}}                    & 13.05 & $\textemdash$ & $\textemdash$ & $\textemdash$ & $\textemdash$ & $\textemdash$                    & $\textemdash$ & $\textemdash$ & $\textemdash$ \\
		SeRankDet  \cite{SeRankDet}               & TGRS'24 & 108.89M                        & 568.74G & \textbf{\textcolor{cyan}{73.66}}                     & \textbf{\textcolor{orange}{69.11}}  & 94.28                     & 8.37  & \textbf{\textcolor{cyan}{75.49}}  & \textbf{\textcolor{green}{71.08}}  & \textbf{\textcolor{orange}{98.97}}  & \textbf{\textcolor{orange}{18.9}}   & 94.28                     & 93.69  & \textbf{\textcolor{cyan}{99.77}}  & 2.03   \\
		IRPruneDet \cite{zhang2024irprunedet}                   & AAAI'24  & 0.1802M                         & $\textemdash$  & 64.54                    & 62.71  & 91.74                     & 16.04 &  $\textemdash$ &  $\textemdash$ &  $\textemdash$ &  $\textemdash$ & $\textemdash$ &  $\textemdash$ & $\textemdash$ & $\textemdash$ \\
		TCI-Former \cite{chen2024tci}                  & AAAI'24  & 3.66M                         & 5.83G  & 70.14                    & 67.69  & \textbf{\textcolor{green}{96.31}}                     & 14.81 &  $\textemdash$ &  $\textemdash$ &  $\textemdash$ &  $\textemdash$ & $\textemdash$ &  $\textemdash$ & $\textemdash$ & $\textemdash$ \\
		MSHNet  \cite{MSHNet}                  & CVPR'24 & 4.07M                          & 24.43G  & 67.87 & 61.70   & 92.86 & 8.88  & $\textemdash$ & $\textemdash$ & $\textemdash$ & $\textemdash$ & 80.55 & $\textemdash$ & 97.99  & 11.77  \\
		HintHCFNet  \cite{10764792}                  & TGRS'24 & $\textemdash$  & $\textemdash$  & 62.24 & 58.87   & 90.17 & 17.12  & $\textemdash$ & $\textemdash$ & $\textemdash$ & $\textemdash$ & 93.29 & 93.67 & 98.93  & \textbf{\textcolor{orange}{0.67}}  \\
		$\text{Mrf}^3\text{Net}$ \cite{MRF3Net} & TGRS'24 & 0.538M                         & 33.2G   & 67.79                     & \textbf{\textcolor{green}{68.74}}  & 95.28                     & 14.5  & $\textemdash$ & $\textemdash$ & $\textemdash$ & $\textemdash$ & 95.21                     & \textbf{\textcolor{green}{95.23}}  & 99.36  & 1.86   \\
		SCTransNet \cite{SCTransNet}               & TGRS'24 & 11.19M                         & 67.4G   & 68.03                     & 68.15  & 93.27                     & 10.74 & $\textemdash$ & $\textemdash$ & $\textemdash$ & $\textemdash$ & 94.09                     & 94.38  & 98.62  & 4.29   \\
		MLP-Net \cite{10793117}                  & TGRS'24  & 8.03M                         & 8.714G  & 56.67                    & 57.16  & 92.01                     & 13.12 &  $\textemdash$ &  $\textemdash$ &  $\textemdash$ &  $\textemdash$ & 90.44 &  92.28 & 99.73 & 1.16 \\
		MDAFNet  \cite{li2024moderately}                 & TGRS'24  & 10.07M                         & 100.46G  & $\textemdash$                    & 64.25  & 76.66                     & 6.11 &  $\textemdash$ &  $\textemdash$ &  $\textemdash$ &  $\textemdash$ & $\textemdash$ &  93.42 & 95.58 & 1.89 \\
		DCFR-Net \cite{fan2024diffusion}                 & TGRS'24  & $\textemdash$                         & $\textemdash$ & 65.41                    & 65.45  & 93.60                     & 7.34 &  $\textemdash$ &  $\textemdash$ &  $\textemdash$ &  $\textemdash$ & 86.93 &  86.92 & 98.26 & 2.48 \\
		CSRNet \cite{CSRNet}       & TIP'24  & 0.4045M                        & 121G    & 65.87                     & 66.70   & \textbf{\textcolor{cyan}{98.16}}                     & 12.08 & $\textemdash$ & $\textemdash$ & $\textemdash$ & $\textemdash$ & $\textemdash$                    & $\textemdash$ & $\textemdash$ & $\textemdash$ \\
		OIPF-SCT \cite{ma2025oipf}                  & TAES'25  & 22.37M                         & 80.97G  & 66.60                     & 66.31  & 94.95                     & 12.1 &  $\textemdash$ &  $\textemdash$ &  $\textemdash$ &  $\textemdash$ & \textbf{\textcolor{orange}{95.43}} &  \textbf{\textcolor{orange}{95.53}} & 99.26 & \textbf{\textcolor{green}{0.90}} \\
		CFD-Net \cite{10949663}                  & TNNLS'25  & 1.05M                         & 110.37G  & 69.97                    & $\textemdash$  & 95.96                     & 14.21 &  $\textemdash$ &  $\textemdash$ &  $\textemdash$ &  $\textemdash$ & $\textemdash$ &  $\textemdash$ & $\textemdash$ & $\textemdash$ \\
		L2SKNet \cite{10813615}                  & TGRS'25  & 0.899M                         & 6.89G  & 67.81                    & $\textemdash$  & 90.24                     & 17.46 &  74.00 &  $\textemdash$ &  \textbf{\textcolor{cyan}{99.17}} &  54.90 & 93.58 &  $\textemdash$ & 97.57 & 5.33 \\
	 	IRMamba \cite{zhang2025irmamba}                  & AAAI'25  & 10.51M                         & $\textemdash$  & 70.04                    & $\textemdash$  & 95.81                     & 5.92 &  $\textemdash$ &  $\textemdash$ &  $\textemdash$ &  $\textemdash$ & 95.18 &  $\textemdash$ & 99.26 & 1.309 \\
	 	PConv \cite{yang2025pinwheel}                  & AAAI'25  & $\textemdash$                         & $\textemdash$  & 67.45                   & $\textemdash$  & 92.20                     & 10.70 &  $\textemdash$ &  $\textemdash$ &  $\textemdash$ &  $\textemdash$ & $\textemdash$ &  $\textemdash$ & $\textemdash$ & $\textemdash$ \\
	 	MMLNet \cite{li2025multi}                   & TGRS'25  & $\textemdash$                         & $\textemdash$  & 67.21                   & $\textemdash$  & 94.28                     & 14.00 &  $\textemdash$ &  $\textemdash$ &  $\textemdash$ &  $\textemdash$ & 81.81 &  $\textemdash$ & 98.43 & 11.77 \\
	 	HDNet  \cite{11017756}                 & TGRS'25  & $\textemdash$                         & $\textemdash$  & 70.26                   & $\textemdash$  & 94.56                     & \textbf{\textcolor{orange}{4.33}} &  $\textemdash$ &  $\textemdash$ &  $\textemdash$ &  $\textemdash$ & 85.17 &  $\textemdash$ & 98.52 & 2.78 \\
		\hline
		\multicolumn{16}{l}{\textit{Deep Unfolding-Based Methods}}  \\
		\hline
		RPCANet  \cite{RPCANet}                 & WACV'24 & 0.68M                          & 179.74G  & 63.21                     & 64.27 & 88.31                     & 43.9  & 72.54  & $\textemdash$ & 98.21  & 34.14  & 89.31                     & 89.03 & 97.14  & 28.7   \\
		\hline
		\multicolumn{16}{l}{\textit{Visual Language Model: SAM + CLIP}}   \\
		\hline
		SAIST \cite{zhang2025saist}                & CVPR'25 & $\textemdash$            & $\textemdash$  & \textbf{\textcolor{green}{72.14}}                     & $\textemdash$  & 96.18                     & \textbf{\textcolor{green}{4.76}}  & $\textemdash$  & $\textemdash$ & $\textemdash$  & $\textemdash$  & \textbf{\textcolor{green}{95.23}}                     & $\textemdash$ & 99.28  & 1.31   \\ 
		\noalign{\hrule height 1pt} \rowcolor[rgb]{0.9,0.9,0.9}
		\multicolumn{16}{l}{\textit{Our proposed method}}  \\
		\hline \rowcolor[rgb]{0.9,0.9,0.9}
		LRRNet (Ours) & $\textemdash$ & 3.45M & 45.5G & \textbf{\textcolor{orange}{72.35}} & \textbf{\textcolor{cyan}{69.51}} & 95.96 & \textbf{\textcolor{cyan}{3.36}} & \textbf{\textcolor{orange}{75.39}} & \textbf{\textcolor{cyan}{71.58}} & \textbf{\textcolor{cyan}{99.17}} & \textbf{\textcolor{cyan}{3.15}} & \textbf{\textcolor{cyan}{96.37}} & \textbf{\textcolor{cyan}{96.72}} & \textbf{\textcolor{green}{99.34}} & \textbf{\textcolor{cyan}{0.20}} \\
		\noalign{\hrule height 1pt}
	\end{tabular}
\end{table*}

\begin{table}[h]
	\renewcommand\arraystretch{1.2}
	\footnotesize
	\centering
	\caption{Performance comparison of different methods on the NoisySIRST dataset \cite{SeRankDet}. The metrics considered include IoU ($10^{-2}$), $\text{nIoU}$ ($10^{-2}$).}
	\label{tab:noise}
	\setlength{\tabcolsep}{3.5pt}
	\begin{tabular}{l|c|cc|cc|cc}
		\noalign{\hrule height 1pt}
		\multirow{2}{*}{Method} & \multirow{2}{*}{Param}  & \multicolumn{2}{c|}{$\sigma_n=10$}  & \multicolumn{2}{c|}{$\sigma_n=20$}   & \multicolumn{2}{c}{$\sigma_n=30$} \\ \multicolumn{1}{l|}{} & \multicolumn{1}{l|}{}
		&  IoU $\uparrow$ &  nIoU  $\uparrow$ & IoU $\uparrow$ &  nIoU $\uparrow$ & IoU $\uparrow$ &  nIoU $\uparrow$ \\
		\noalign{\hrule height 1pt}
		
		ACM \cite{Dai_2021_WACV} & 0.5M & 73.71 & 71.70 & 69.66 & 68.24 & 66.95 & 65.38 \\
		AGPCNet \cite{AGPCNet} & 12M & 71.41 & 70.96 & 71.30 & 69.90 & 68.05 & 67.28 \\
		DNANet \cite{DNANet} & 4.6M & 71.44 & 72.57 & 70.53 & 68.14 & 68.50  & 66.42 \\
		UIUNet \cite{UIUNet} & 50M & 77.01 & 74.45 & 73.36 & 70.44 & 69.16 & 69.17 \\
		RDIAN \cite{RDIAN}  & 0.1M & 70.86 &  71.57 & 70.82 & 68.40 & 66.09 & 66.01 \\  
		MTUNet \cite{MTUNet} & 4.0M & 75.42 & 73.11 & 71.27 & 69.59 & 66.40 & 66.94 \\
		ABC \cite{ABC} & 73M & 77.33  & 74.97  &  73.01 & 70.15  &  69.24 & 68.93 \\
		SeRankDet \cite{SeRankDet} &108M & \underline{77.59} & \textbf{75.30} & \underline{73.49} & \underline{71.94} & \underline{69.68} & \textbf{69.68} \\  \rowcolor[rgb]{0.9,0.9,0.9}
		LRRNet (Ours) &3.4M & \textbf{77.78} & \underline{75.24} & \textbf{74.11} & \textbf{72.23} & \textbf{69.87} & \underline{69.43} \\
		\noalign{\hrule height 1pt}
	\end{tabular} \label{Tab:Noisy}
\end{table}

We compare the proposed method with related model-driven and data-driven methods. The model-driven methods include Low-rank and Sparse Decomposition, while the data-driven methods encompass Deep Learning Methods, Deep Unfolding-Based Methods, and Visual Language Model. Specifically, the compared algorithms are as follows:
\begin{itemize}
	\item Low-rank and Sparse Decomposition: IPI \cite{gao2013infrared}, NIPPS \cite{dai2017non}, RIPT \cite{dai2017reweighted}, NRAM \cite{zhang2018infrared}, NOLC \cite{zhang2019infrared1}, PSTNN \cite{zhang2019infrared}.
	\item Deep Learning Methods: MDvsFA \cite{wang2019miss}, ACM  \cite{Dai_2021_WACV}, ALCNet  \cite{ALCNet}, FC3-Net  \cite{FC3-Net}, DNANet \cite{DNANet},  ISNet \cite{ISNet}, AGPCNet \cite{AGPCNet}, UIUNet  \cite{UIUNet}, RDIAN  \cite{RDIAN}, MTUNet \cite{MTUNet}, 	RepISD-Net \cite{wu2023repisd}, SRNet  \cite{SRNet}, SeRankDet  \cite{SeRankDet}, IRPruneDet \cite{zhang2024irprunedet}, TCI-Former \cite{chen2024tci}, MSHNet  \cite{MSHNet},	HintHCFNet  \cite{10764792}, $\text{Mrf}^3\text{Net}$ \cite{MRF3Net}, SCTransNet \cite{SCTransNet}, MLP-Net \cite{10793117}, MDAFNet  \cite{li2024moderately}, DCFR-Net \cite{fan2024diffusion}, CSRNet \cite{CSRNet}, OIPF-SCT \cite{ma2025oipf}, CFD-Net \cite{10949663}, L2SKNet \cite{10813615}, IRMamba \cite{zhang2025irmamba}, PConv \cite{yang2025pinwheel}, MMLNet \cite{li2025multi}, HDNet  \cite{11017756}.
	\item Deep Unfolding-Based Methods: RPCANet \cite{RPCANet}.
	\item Visual Language Model: SAIST \cite{zhang2025saist}.
\end{itemize}

\subsubsection{Quantitative Evaluation}

The performance of various methods across the three public datasets is summarized in Tab.~\ref{tab:sota}. As shown, LRRNet consistently ranks among the top performers in terms of IoU, nIoU, $P_d$, and $F_a$ metrics. It achieves the highest nIoU, which is particularly indicative of segmentation accuracy for small targets, and maintains the lowest $F_a$ across all datasets. Although LRRNet’s IoU is slightly lower than that of SeRankDet \cite{SeRankDet} on the IRSTD-1K and SIRSTAUG datasets, it is important to note that SeRankDet attains these results with a model size of 108M parameters. In contrast, LRRNet achieves competitive performance with only 3.4M parameters.
In general, feature learning-based methods often struggle with small targets of varying shapes and sizes. This challenge arises because infrared small targets do not belong to well-defined object categories—distant humans, vehicles, and drones are all uniformly labeled as small targets—making target features inherently ambiguous and difficult to generalize. As a result, approaches that rely on learning explicit target representations tend to exhibit limited robustness. Deep unfolding methods, such as RPCANet \cite{RPCANet}, perform even worse than conventional black-box models due to their reliance on rigid and often unrealistic assumptions. While SAIST \cite{zhang2025saist} introduces improvements by integrating CLIP and SAM, the contextual dependency of small targets in infrared imagery \cite{LCRNet} limits the utility of simple language prompts. Moreover, the texture bias inherent in foundation models further reduces their effectiveness in handling the texture-sparse characteristics of infrared images \cite{zhang2025vision}.

\begin{table}[]
	\renewcommand\arraystretch{1.2}
	\footnotesize
	\centering
	\caption{Time Consumption of the Compared typical Methods (FPS)}
	\label{tab:fps}
	\setlength{\tabcolsep}{3.5pt}
	\begin{tabular}{cccccccc}
		\toprule
		IPI & NIPPS & RIPT & PSTNN & DNANet & ALCNet & RDIAN & LRRNet  \\ \midrule
		0.3 & 0.4   & 1.5  & 4.4   & 4.56   & 11.8   & 40    & 82.34 \\ \bottomrule
	\end{tabular}
\end{table}

It is worth noting that our LRRNet achieves a throughput of 82.34 FPS on an NVIDIA 3080Ti GPU with an input resolution of $512 \times 512$ and a batch size of 1. In contrast, as shown in Tab. \ref{tab:fps}, methods that rely on learning object-specific features often suffer from significantly slower inference speeds \cite{MRF3Net}. For example, under the same conditions, DNANet \cite{DNANet} achieves only 4.56 FPS. Even RDIAN \cite{RDIAN}, which is specifically designed for lightweight architecture and high inference speed, still performs considerably slower than our LRRNet.
This performance gap is primarily due to the use of complex module designs aimed at incorporating small-object information into deeper layers, which substantially increases computational overhead and slows down inference.
Moreover, compared with traditional model-driven methods, LRRNet demonstrates significantly higher FPS. This is because LRRNet avoids time-consuming alternating iterations and instead performs target detection through a single forward pass, fully leveraging the highly parallel computing capabilities of GPUs.

To further assess model robustness, we conduct additional experiments on the NoisySIRST dataset \cite{SeRankDet}, which consists of images degraded by varying levels of white noise to simulate the challenging conditions encountered with low-end infrared cameras. As shown in Tab.~\ref{Tab:Noisy}, there exists a general trend in which model robustness improves with increased parameter count among models that rely on learning target-specific features. Notably, models that perform well on this dataset typically contain more than 50M parameters. In contrast, despite having only 3.4M parameters, LRRNet delivers competitive performance. This robustness stems from its ability to capture the low-rank structure of the background. By effectively suppressing the background and leveraging local smoothness priors, LRRNet distinguishes true targets from noise, thereby maintaining high performance under extreme conditions.

\begin{figure}
	\centering
	\includegraphics[width=\linewidth]{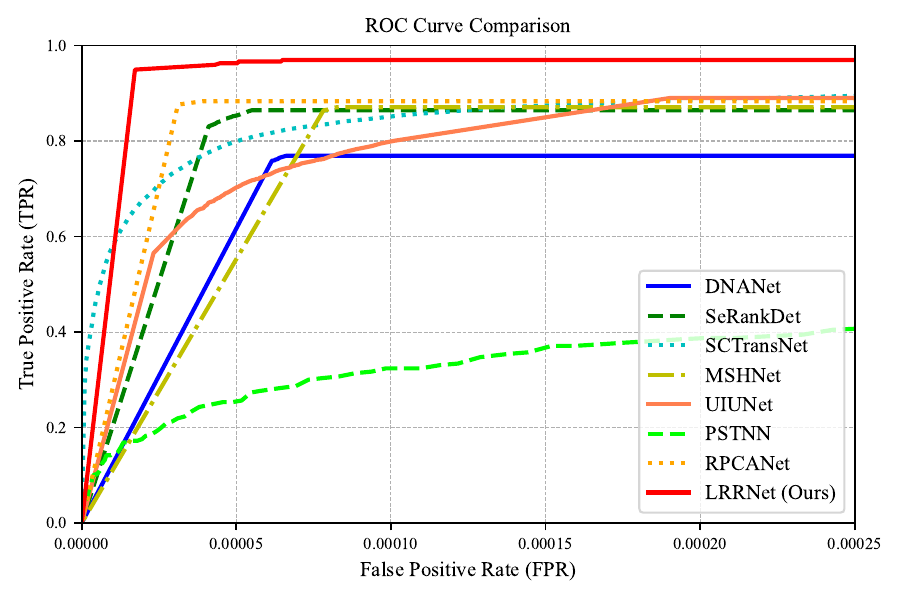}
	\caption{ROC curves of our LRRNet and other approaches on IRSTD-1k \cite{ISNet}.}
	\label{fig:ROC}   
\end{figure}

To further validate this robustness, we present the ROC curves in Fig.~\ref{fig:ROC}, evaluated on the IRSTD-1K dataset, where our method achieves the highest AUC score. Moreover, by leveraging the robust characteristic of the low-rank structure inherent in the background of infrared images, the proposed method achieves an excellent balance between detection rate and false alarm rate.

\subsubsection{Qualitative Evaluation}
\begin{figure}[!t]
	\centering
	\subfloat[]{
		\includegraphics[width=\linewidth]{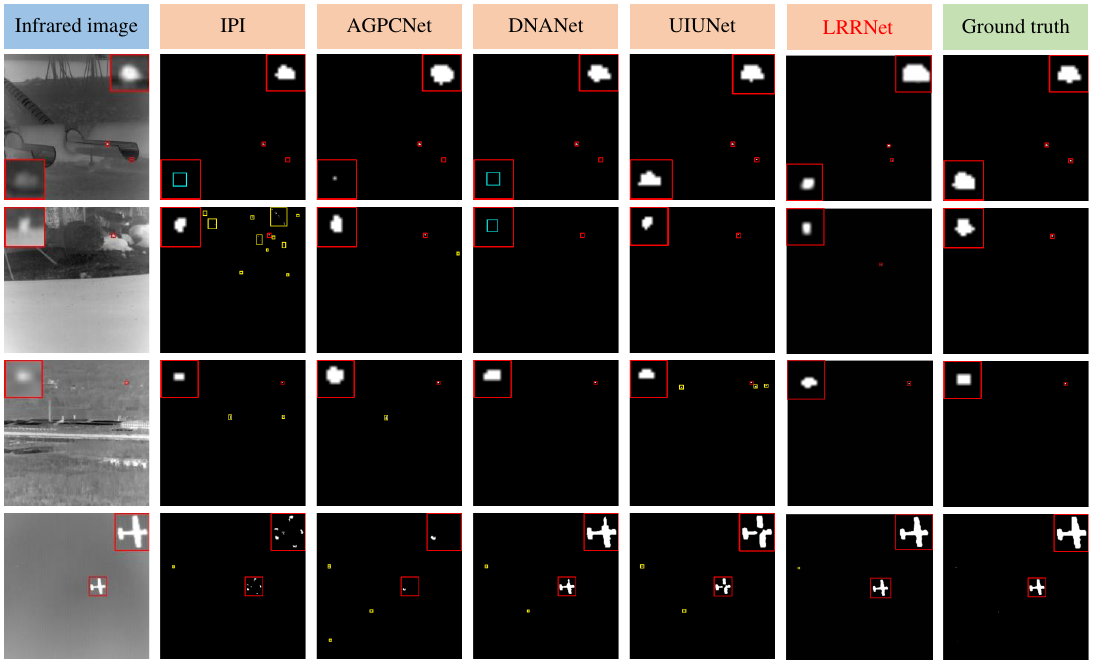}
		\label{fig:visual}
	}
	\hfill
	\subfloat[]{
		\includegraphics[width=\linewidth]{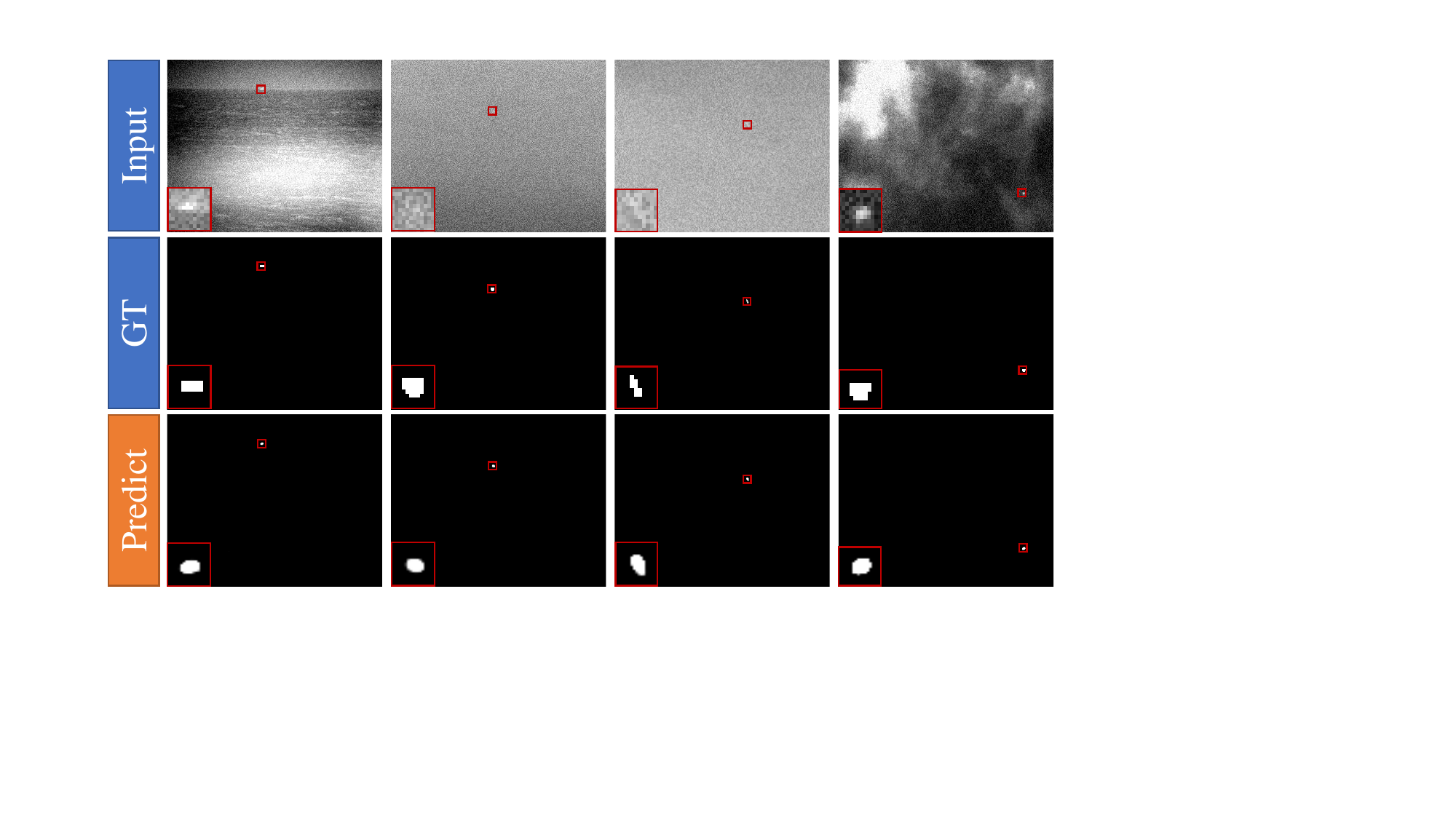}
		\label{fig:visual_noise}
	}
	\caption{Qualitative evaluation for Our LRRNet. (a) Visualization comparison of detection results via different methods on representative images from IRSTD-1k dataset. The red, yellow, and cyan boxes denote correct detections, false alarms, and missed detections, respectively. (b) Visualization of the LRRNet results on the NoisySIRST dataset under Gaussian white noise interference with a variance of 30.}
\end{figure}

The qualitative results in Fig.~\ref{fig:visual} provide a visual comparison of detection performance on representative samples from the IRSTD-1K dataset, where true targets (red boxes), false alarms (yellow boxes), and missed detections (cyan boxes) are highlighted. It is evident that our LRRNet achieves the best overall performance, demonstrating strong adaptability to multi-scale targets, whether they are typical small targets or substantially larger ones. Although some false alarms are present in the visualization results produced by LRRNet, these false positives cannot be reliably determined from single frames alone. Notably, similar false alarms are also observed across all compared methods. Such false alarms are likely more effectively suppressed by leveraging temporal information through multi-frame infrared small target detection approaches, which can exploit spatiotemporal consistency to distinguish true targets from noise and clutter.
To further demonstrate the effectiveness of LRRNet, we present visualization results on the NoisySIRST dataset under Gaussian white noise interference with a variance of 30, as shown in Fig. \ref{fig:visual_noise}. It can be observed that even in cases where the targets are difficult to discern by the human eye, the proposed LRRNet still exhibits remarkable performance. This is attributed to the application of reconstruction loss, which effectively suppresses the interference caused by Gaussian noise. 

\subsection{Ablation Study}
\subsubsection{The Effect of Densely Connection}
To evaluate the effectiveness of the densely connection, we perform an ablation study by replacing its dense skip connections with standard ones, yielding a conventional U-Net. Both variants share the same configuration, including residual blocks, channel dimensions, and a self-attention module at the lowest resolution level \cite{liu2023infrared}. This experiment is conducted on the IRSTD-1K dataset \cite{ISNet} to isolate the impact of dense connectivity.

As shown in Table~\ref{Tab:UNet}, the densely connected U-Net significantly outperforms the baseline: IoU increases by 1.90, nIoU by 1.8, $P_d$ by 1.68, and $F_a$ decreases by 11.29. This improvement stems from the limitation of standard skip connections, which, despite preserving information, suffer from feature confusion in deeper layers \cite{zhou2021deepvit}, hindering accurate background reconstruction. Moreover, deep equilibrium models \cite{bai2019deep} suggest that even identical modules can respond differently to various features. Dense long-range connections facilitate finer-grained feature propagation, enabling better preservation of subtle structures in cluttered scenes.
\begin{table}[]
	\centering
	\caption{Ablation Study on Compression–Reconstruction Module Using the IRSTD-1K Dataset.}
	\begin{tabular}{c|cccc}
		\noalign{\hrule height 1pt}
		\multirow{2}{*}{Variants} & \multicolumn{4}{c}{IRSTD-1K \cite{ISNet}}  \\
		& IoU $\uparrow$   & nIoU$\uparrow$ & $P_d$ (\%) $\uparrow$   & $F_a$ ($ 10^{-6}$) $\downarrow$    \\ 	\noalign{\hrule height 1pt}
		\multicolumn{5}{c}{The effect of Densely Connection}                                   \\ \hline  \rowcolor[rgb]{0.9,0.9,0.9}
		w. Densely Connection    & \textbf{72.35} & \textbf{69.51} & \textbf{95.96} & \textbf{3.36}  \\
		w/o Densely Connection   & 70.45 & 67.71 & 93.94 & 14.65 \\
		\hline 	
		\multicolumn{5}{c}{The effect of long skip connections without attention}                                   \\ \hline  \rowcolor[rgb]{0.9,0.9,0.9}
		w/o attention & \textbf{72.35} & \textbf{69.51} & \textbf{95.96} & \textbf{3.36} \\
		w. CBAM \cite{woo2018cbam} & 68.10 & 65.68 & 91.92 & 25.94 \\
		w. SE \cite{8578843} & 69.98 & 65.82 & 89.56& 15.89 \\
		w. SimAM \cite{pmlr-v139-yang21o} & 68.11 & 67.00 & 92.93 & 10.86 \\
		\hline
		\multicolumn{5}{c}{The effect of self-attention}                                   \\ \hline  \rowcolor[rgb]{0.9,0.9,0.9}
		w. self-attention & \textbf{72.35} & \textbf{69.51} & \textbf{95.96} & \textbf{3.36} \\
		w. TGM \& TRM \cite{chen2020tensor} & 63.31 & 62.58 & 91.73 & 35.55 \\
		w/o self-attention & 70.46 & 69.08 & 92.93 & 15.77 \\
		\hline
		\multicolumn{5}{c}{The effect of the number of self-attention} \\
		\hline  \rowcolor[rgb]{0.9,0.9,0.9}
		Only one & \textbf{72.35} & \textbf{69.51} & \textbf{95.96} & \textbf{3.36} \\
		Two Blocks & 71.57 & 69.22 & 90.91 & 10.19 \\ \hline
		\multicolumn{5}{c}{The effect of the Pre-act ResBlock} \\ \hline  \rowcolor[rgb]{0.9,0.9,0.9}
		w. Pre-act ResBlock & \textbf{72.35} & \textbf{69.51} & \textbf{95.96} & \textbf{3.36} \\
		w. ResBlock & 67.29 & 67.29 & 91.92 & 20.08 \\
		\noalign{\hrule height 1pt}
	\end{tabular} \label{Tab:UNet}
\end{table}

\subsubsection{The Effect of Long Skip Connections Without Attention}
Many existing UNet variants for infrared small target detection incorporate attention mechanisms, such as SE \cite{8578843}, CBAM \cite{woo2018cbam}, or SimAM \cite{pmlr-v139-yang21o}—along long skip connections to enhance feature selection. These modules are effective in highlighting salient targets, particularly when the model focuses on target-specific features \cite{DNANet}.

In contrast, our design intentionally omits attention mechanisms along the skip paths. As shown in Table~\ref{Tab:UNet}, this simple yet effective choice yields superior performance with statistically significant gains. We attribute this to the tendency of attention modules to overemphasize dominant structures, which, while beneficial for target enhancement, can hinder accurate low-rank background reconstruction. By avoiding such interference, our approach enables more precise modeling of the background and clearer separation between targets and clutter.

\subsubsection{The Effect of Self-Attention}
The self-attention mechanism \cite{liu2023infrared} is essential for capturing long-range dependencies, making it particularly effective for modeling the low-rank structure of background regions. Placing self-attention at the lowest-resolution feature map also reduces computational cost due to the smaller spatial size.

As shown in Table~\ref{Tab:UNet}, introducing a single self-attention block significantly improves performance. However, adding multiple blocks leads to degradation. This can be attributed to the lack of inductive bias in self-attention \cite{liu2023infrared}; when overused, it behaves like deep clustering—useful for compression but detrimental to reconstruction fidelity. Thus, a restrained application of self-attention is key to balancing compression efficiency with accurate reconstruction. Furthermore, we attempted to replace the self-attention mechanism with a tensor-based alternative, namely the Tensor Generation Module (TGM) and Tensor Reconstruction Module (TRM) proposed in \cite{chen2020tensor}, which perform explicit CP tensor decomposition. However, this substitution resulted in a significant drop in performance. We attribute this to the fact that explicit low-rank constraints in infrared imagery are more appropriately applied at the patch-image level rather than the image level. Direct decomposition in the full image domain imposes overly rigid constraints that hinder effective feature representation and reconstruction.

\begin{table}[]
	\centering
	\caption{Ablation Study on Subtraction Module Using the IRSTD-1K Dataset.}
	\begin{tabular}{ccccc}
		\noalign{\hrule height 1pt}
		\multicolumn{1}{c|}{\multirow{2}{*}{Subtraction   Module Variants}} & \multicolumn{4}{c}{IRSTD-1K}  \\
		\multicolumn{1}{c|}{}                                      & IoU $\uparrow$   & nIoU$\uparrow$ & $P_d$ $\uparrow$   & $F_a$ $\downarrow$    \\ \noalign{\hrule height 1pt}
		\multicolumn{5}{c}{The effect of subtraction strategy}                                   \\ \hline  \rowcolor[rgb]{0.9,0.9,0.9}
		\multicolumn{1}{c|}{Subtraction}                           & \textbf{72.35} & \textbf{69.51} & \textbf{95.96} & \textbf{3.36}  \\
		\multicolumn{1}{c|}{Addition}                              & 68.82 & 67.78 & 92.93 & 25.64 \\ 
		\hline
		\multicolumn{5}{c}{The effect of the   reconstruction strategy}                            \\ \hline  \rowcolor[rgb]{0.9,0.9,0.9}
		\multicolumn{1}{c|}{w. reconstruction strategy}             & \textbf{72.35} & \textbf{69.51} & \textbf{95.96} & \textbf{3.36}  \\
		\multicolumn{1}{c|}{w/o reconstruction strategy}           & 70.84 & 67.23 & 91.92 & 12.53 \\\hline
		\multicolumn{5}{c}{The effect of   ResBlock}                                                \\ \hline  \rowcolor[rgb]{0.9,0.9,0.9}
		\multicolumn{1}{c|}{Pre-act ResBlock}                              & \textbf{72.35} & \textbf{69.51} & \textbf{95.96} & \textbf{3.36}   \\
		\multicolumn{1}{c|}{Simple $\text{Conv}_{3\times3}$}                        & 67.98 & 66.59 & 91.92 & 19.32 \\  \hline
		\multicolumn{5}{c}{The effect of the number of ResBlocks}                                \\ \hline \rowcolor[rgb]{0.9,0.9,0.9}
		\multicolumn{1}{c|}{Only one}                               & \textbf{72.35} & \textbf{69.51} & \textbf{95.96} & \textbf{3.36}  \\
		\multicolumn{1}{c|}{Two blocks}                            & 70.92 & 68.13 & 93.27 & 14.97 \\ \hline
		\multicolumn{5}{c}{The effect of the Pre-act ResBlock} \\ \hline \rowcolor[rgb]{0.9,0.9,0.9}
		w. Pre-act ResBlock & \textbf{72.35} & \textbf{69.51} & \textbf{95.96} & \textbf{3.36} \\
		w. Conventional ResBlock \cite{huang2023revisiting} & 66.33 & 67.95 & 94.28 & 64.20 \\
		\noalign{\hrule height 1pt}
	\end{tabular} \label{Tab:Sub}
\end{table}

\subsubsection{The Effect of Subtraction Strategy} The subtraction strategy separates target-related information by subtracting the reconstructed background feature map from the corresponding feature map of the original image. If replaced with addition, the model effectively reverts to a conventional target feature learning approach. As shown in Tab.~\ref{Tab:Sub}, the subtraction strategy significantly enhances model performance.

\subsubsection{The Effect of Reconstruction Strategy} The reconstruction strategy is also a key component of LRRNet. On one hand, it guides the model to reconstruct the background; on the other, the use of MSE loss enhances the model's robustness to Gaussian noise. As shown in Tab.~\ref{Tab:Sub}, the reconstruction strategy has a significant impact on overall performance.

\subsubsection{The Effect of ResBlock} The ResBlock plays an important role in effectively suppressing noise in the input image \cite{7839189}. As shown in Tab.~\ref{Tab:Sub}, simply replacing it with a $3 \times 3$ convolution results in a significant performance drop. However, increasing the number of ResBlocks excessively can also be detrimental. Although short skip connections help preserve features, combining them through nonlinear operations may introduce feature confusion. Given the small size of the targets, even minor disturbances can lead to noticeable performance degradation.

\subsubsection{The Effect of Pre-act ResBlock} The structure of the ResBlock also has a considerable impact on performance. In particular, the Pre-activation ResBlock demonstrates stronger robustness to noise compared to other variants \cite{huang2023revisiting}, which is \textbf{critical} for both the compression–reconstruction and subtraction processes. As shown in Tab.~\ref{Tab:UNet} and \ref{Tab:Sub}, modifying the ResBlock structure in either the compression–reconstruction module or the subtraction module leads to significant performance changes.

\begin{table}[]
	\centering
	\caption{Ablation Study on Hyperparameter Using the IRSTD-1K Dataset.}
	\scalebox{1.05}{\begin{tabular}{ccccc}
			\noalign{\hrule height 1pt}
			\multicolumn{1}{c|}{\multirow{2}{*}{Hyperparameter Discussion}} & \multicolumn{4}{c}{IRSTD-1K}  \\
			\multicolumn{1}{c|}{}                                      & IoU $\uparrow$   & nIoU$\uparrow$ & $P_d$ $\uparrow$   & $F_a$ $\downarrow$    \\ \noalign{\hrule height 1pt}
			\multicolumn{5}{c}{The effect of the number of $\lambda$ in loss function}                                \\ \hline
			\multicolumn{1}{c|}{$\lambda$ = 0.0} & 70.84 & 67.23 & 91.92 & 12.53 \\ \rowcolor[rgb]{0.9,0.9,0.9}
			\multicolumn{1}{c|}{$\lambda$ = 0.1} & \textbf{72.35} & \textbf{69.51} & \textbf{95.96} & \textbf{3.36} \\
			\multicolumn{1}{c|}{$\lambda$ = 0.2} & 72.05 & 68.43 & 92.26 & 10.36 \\
			\multicolumn{1}{c|}{$\lambda$ = 0.3} & 70.86 & 69.17 & 90.57 & 8.67 \\
			\multicolumn{1}{c|}{$\lambda$ = 0.4} & 71.41 & 69.20 & 92.26 & 13.97 \\
			\noalign{\hrule height 1pt}
	\end{tabular}} \label{Tab:Lambda}
\end{table}

\subsubsection{Hyperparameter Discussion} The key hyperparameter in our LRRNet is the weight $\lambda$ in the loss function, which balances the optimization contributions from the segmentation and reconstruction branches. The impact of different $\lambda$ values on the final performance is shown in Tab. \ref{Tab:Lambda}. When $\lambda = 0$, the model lacks constraints from the compression-reconstruction process, leading to degraded performance. On the other hand, a large $\lambda$ causes the reconstruction branch to dominate the optimization. An interesting observation during training is that when $\lambda$ is large, the segmentation loss remains nearly stagnant in the early stages and only begins to decrease when the reconstruction loss has been sufficiently minimized.

\begin{table}[]
	\centering
	\caption{Ablation Study on Model Size Using the IRSTD-1K Dataset.}
	\scalebox{1.05}{\begin{tabular}{ccccc}
			\noalign{\hrule height 1pt}
			\multicolumn{1}{c|}{\multirow{2}{*}{Model Size Discussion}} & \multicolumn{4}{c}{IRSTD-1K}  \\
			\multicolumn{1}{c|}{}                                      & IoU $\uparrow$   & nIoU$\uparrow$ & $P_d$ $\uparrow$   & $F_a$ $\downarrow$    \\ \noalign{\hrule height 1pt}
			\multicolumn{5}{c}{The effect of the number of Channels}                                \\ \hline
			\multicolumn{1}{c|}{[8, 16, 32, 64]} & 69.03 & 68.37 & 94.28 & 30.10 \\ \rowcolor[rgb]{0.9,0.9,0.9}
			\multicolumn{1}{c|}{[16, 32, 64, 128]} &\textbf{72.35} &\textbf{69.51} &\textbf{95.96} &\textbf{3.36} \\
			\multicolumn{1}{c|}{[32, 64, 128, 256]} & 67.79 & 68.20 & 93.27 & 21.43 \\
			\hline
			\multicolumn{5}{c}{The effect of the number of Stages} \\ \hline 
			\multicolumn{1}{c|}{Stage = 3} & \multicolumn{4}{c}{Out Of Memory} \\ \rowcolor[rgb]{0.9,0.9,0.9}
			\multicolumn{1}{c|}{Stage = 4} & \textbf{72.35} & \textbf{69.51} & \textbf{95.96} & \textbf{3.36} \\
			\multicolumn{1}{c|}{Stage = 5} & 71.38 & 68.21 & 94.61 & 20.35 \\
			\hline
			\multicolumn{5}{c}{The effect of the number of ResBlocks in DB} \\ \hline 
			\multicolumn{1}{c|}{number = 1} & 69.73 & 68.30 & 94.91 & 35.51 \\ \rowcolor[rgb]{0.9,0.9,0.9}
			\multicolumn{1}{c|}{number = 2} & \textbf{72.35} & \textbf{69.51} & \textbf{95.96} & \textbf{3.36} \\
			\multicolumn{1}{c|}{number = 3} & 69.23 & 68.55 & 93.27 & 49.71 \\
			\noalign{\hrule height 1pt}
	\end{tabular}} \label{Tab:Size}
\end{table}

\subsubsection{Model Size Discussion} Finally, we investigate the impact of architectural scale on model performance, focusing on the optimal configuration of channels, stages, and the number of ResBlocks per stage. As shown in Tab. \ref{Tab:Size}, performance degrades when the number of channels is too small, as insufficient filters hinder the model's ability to capture diverse image features. Conversely, blindly increasing the number of channels may also impair performance, as the additional capacity can lead to feature redundancy and diluted useful information.
In terms of stage design, reducing the number of stages leads to out-of-memory (OOM) errors during training. This is because the self-attention module, applied at the lowest-resolution feature map, incurs a quadratic computational cost, which becomes prohibitive as the resolution increases. On the other hand, excessively increasing the number of stages results in over-parameterization, making the model prone to overfitting and degrading generalization.
As illustrated in Fig. \ref{fig:arch}, due to the densely connected design of LRRNet, each decoder stage contains one more ResBlock than its corresponding encoder stage. For clarity, Tab. \ref{Tab:Size} only reports the number of ResBlocks in each encoder stage under different configurations. It can be observed that the model achieves the best performance when each encoder stage contains two ResBlocks.

\subsection{Limitations}
The limitations of our method can be summarized in the following two aspects:
\begin{figure}
	\centering
	\includegraphics[width=\linewidth]{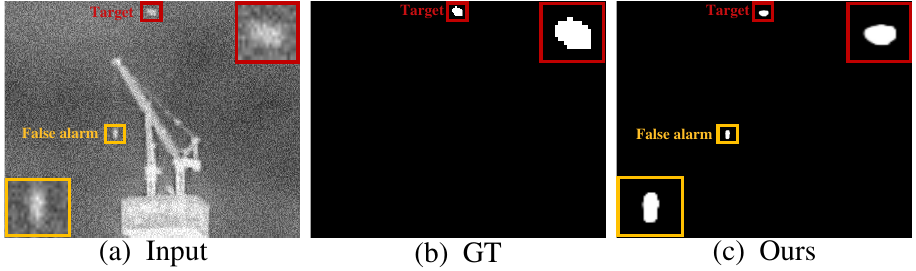}
	\caption{Typical failure cases. The correlation between false alarm sources and the surrounding background is significantly diminished under strong noise interference.}
	\label{fig:Failed}
\end{figure}

First, a representative failure case is depicted in Fig.~\ref{fig:Failed}. When subjected to strong Gaussian white noise with a variance of 30, the correlation at the feature level between false alarm sources and the surrounding background becomes significantly attenuated. As a result, the proposed compress–reconstruct–subtract paradigm is unable to effectively distinguish true targets from false alarms. In such challenging scenarios, incorporating target-specific feature learning may provide enhanced discrimination capability. We intend to explore the integration of target-aware representations into our current framework in future work.

Second, most state-of-the-art infrared small target detection approaches leverage deep supervision~\cite{UIUNet} to achieve high detection accuracy. However, traditional deep supervision schemes are not directly applicable to our LRRNet due to the lack of explicit constraints on intermediate outputs across different stages. Inspired by DNANet~\cite{DNANet}, a potential improvement is to construct a hierarchical feature pyramid by aggregating multi-scale outputs from each encoder stage within the compress–reconstruct module, followed by feature fusion to produce a more refined background representation. We leave this direction for future investigation.

\section{Conclusion}\label{Section:Conclusion}
In this paper, we proposed LRRNet, an end-to-end network that directly models low-rank background priors in the image domain without relying on patch decomposition. By leveraging a compress–reconstruct–subtract paradigm, LRRNet effectively preserves small target features while reducing model complexity. Extensive experiments demonstrate that our method achieves superior accuracy, efficiency, and robustness compared to state-of-the-art approaches. This work highlights the promising potential of integrating interpretable low-rank priors for practical infrared small target detection. In future work, we intend to explore tailored solutions addressing the two aforementioned limitations within this paradigm, thereby further enhancing the framework’s capabilities.

\bibliographystyle{IEEEtran}
\bibliography{reference}

\newpage

\vfill

\end{document}